\documentclass[twoside]{article}

 \usepackage[accepted]{aistats2019}
\usepackage[round]{natbib}
\usepackage{mathtools}
\usepackage{multirow}

\usepackage{amsfonts}
\usepackage{fancyhdr}
\usepackage{algorithm}
\usepackage[noend]{algpseudocode}

\usepackage{comment}
\usepackage{amsmath}
\usepackage{changepage}
\usepackage{graphicx,amsmath,amssymb,fullpage, amsfonts,bbm,bbold}
\usepackage{amssymb}
\usepackage{graphicx}%
\usepackage{tikz}

\usetikzlibrary{arrows}
\usetikzlibrary{positioning}
\usepackage{booktabs}
\newdimen\nodeDist
\usepackage{array}
\usepackage[flushleft]{threeparttable}

\DeclareMathOperator{\iid}{\stackrel{\mbox{\tiny iid} }{\sim}}

\newcommand{\x}{\mathrm{x}}
\newcommand{\w}{\mathrm{w}}
\newcommand{\y}{\mathrm{y}}
\newcommand{\res}{\mathrm{r}}
\newcommand{\N}{\mbox{{\small\textsc{N}}}}
\newcommand{\X}{\mathbf{X}}

\newdimen\nodeDist
\begin{document}

%

%

\twocolumn[

\aistatstitle{XBART: Accelerated Bayesian Additive Regression Trees}

\aistatsauthor{ Jingyu He \And Saar Yalov \And  P. Richard Hahn}

\aistatsaddress{ University of Chicago \And  Arizona State University \And Arizona State University }  ]




\begin{abstract}
Bayesian additive regression trees (BART) \citep{chipman2010bart} is a powerful predictive model that often outperforms alternative models at out-of-sample prediction. BART is especially well-suited to settings with unstructured predictor variables and substantial sources of unmeasured variation as is typical in the social, behavioral and health sciences. This paper develops a modified version of BART that is amenable to fast posterior estimation. We present a stochastic hill climbing algorithm that matches the remarkable predictive accuracy of previous BART implementations, but is many times faster and less memory intensive.  Simulation studies show that the new method is comparable in computation time and 
more accurate at function estimation than both random forests and gradient boosting.

\end{abstract}

\section{INTRODUCTION}
Tree-based regression methods --- CART \citep{breiman1984classification}, random forests \citep{breiman2001random}, and gradient boosting \citep{breiman1997arcing, friedman2001greedy,friedman2002stochastic} --- are highly successful and widely used for supervised learning. Bayesian additive regression trees  --- or BART ---  is a closely related but less well-known method that often achieves superior prediction/estimation accuracy. The ``Bayesian CART'' (single-tree) model was introduced in \cite{chipman1998bayesian} and the BART model first appeared in \cite{chipman2010bart}, although software was publicly available as early as 2006. Contrary to common perception, BART is not merely a  version of random forests or boosted regression trees in which prior distributions have been placed over model parameters. Instead, the Bayesian perspective leads to a fundamentally new tree growing criterion and algorithm, which yields a number of practical advantages --- robustness to the choice of user-selected tuning parameters, more accurate predictions, and a natural Bayesian measure of uncertainty. 

Despite these virtues, BART's wider adoption has been slowed by its more severe computational demands relative to alternatives, owing to its reliance on a random walk Metropolis-Hastings Markov chain Monte Carlo (MCMC) approach. The current fastest implementation, the {\tt R} package {\tt dbarts}, takes orders of magnitude longer than the widely-used {\tt R} package {\tt xgboost}, for example. This paper develops a variant of BART that is amenable to fast posterior estimation, making it almost as fast as {\tt xgboost} (after cross-validating), while still retaining BART's hyperparameter robustness and remarkable predictive accuracy.

First, we describe the BART model to motivate our computational innovations. We derive the BART model's tree-growing criterion, which is notably different than the traditional sum-of-squares criterion used by other methods. We then describe the new algorithm accelerated Bayesian additive regression trees heuristic (XBART) and illustrate its impact on fast, accurate statistical prediction.  Specifically, we compare the new method's performance to random forests, boosted regression trees, neural networks as well as the standard MCMC implementations of BART.

\vspace{-3mm}
\section{BART IN DETAIL}
\subsection{The Model: Likelihood and Prior}
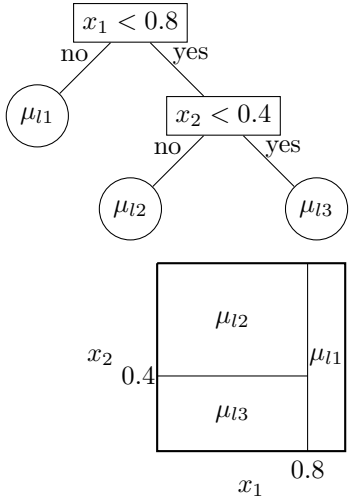
\begin{figure}[h]
\begin{center}
\begin{tikzpicture}[
  scale=0.7,
    node/.style={%
      draw,
      rectangle,
    },
    node2/.style={%
      draw,
      circle,
    },
  ]

    \node [node] (A) {$x_1<0.8$};
    \path (A) ++(-135:\nodeDist) node [node2] (B) {$\mu_{l1}$};
    \path (A) ++(-45:\nodeDist) node [node] (C) {$x_2<0.4$};
    \path (C) ++(-135:\nodeDist) node [node2] (D) {$\mu_{l2}$};
    \path (C) ++(-45:\nodeDist) node [node2] (E) {$\mu_{l3}$};

    \draw (A) -- (B) node [left,pos=0.25] {no}(A);
    \draw (A) -- (C) node [right,pos=0.25] {yes}(A);
    \draw (C) -- (D) node [left,pos=0.25] {no}(A);
    \draw (C) -- (E) node [right,pos=0.25] {yes}(A);
\end{tikzpicture}
\hspace{0.1\linewidth}
\vspace{0.1in}

\begin{tikzpicture}[scale=2.5]
\draw [thick, -] (0,1) -- (0,0) -- (1,0) -- (1,1)--(0,1);
\draw [thin, -] (0.8, 1) -- (0.8, 0);
\draw [thin, -] (0.0, 0.4) -- (0.8, 0.4);
\node at (-0.1,0.4) {0.4};
\node at (0.8,-0.1) {0.8};
\node at (0.5,-0.2) {$x_1$};
\node at (-0.3,0.5) {$x_2$};
\node at (0.9,0.5) {$\mu_{l1}$};
\node at (0.4,0.7) {$\mu_{l2}$};
\node at (0.4,0.2) {$\mu_{l3}$};
\end{tikzpicture}
\end{center}
\caption{(Top) An example binary tree, with internal nodes labelled by their splitting rules and terminal nodes labelled with the corresponding parameters $\mu_{lb}$. (Bottom) The corresponding partition of the sample space and the step function.}
\label{fig:treestep}
\end{figure}

The BART model is an additive error mean regression model 
\begin{equation}\label{additive}
y_i = f(\x_i) + \epsilon_i
\end{equation}
where the $\epsilon_i$ are assumed to be independent mean zero Gaussians and $f(\cdot)$ is an unknown function. The BART prior represents the unknown function $f(\x)$ as a sum of many piecewise constant binary regression trees: 
\begin{equation}\label{forest}
f(\x)=\sum_{l=1}^L g_l(\x, T_l, \mu_l)
\end{equation}
where $T_l$ denotes a regression tree and $\mu_l$ denotes a vector of scalar means associated to the leafs nodes of $T_l$. Each tree $T_l,\;1\leq l\leq L$, consists of a set of internal decision nodes which define a partition of the covariate space (say $\mathcal{A}_1,\dots,\mathcal{A}_{B(l)}$), as well as a set of terminal nodes or leaves corresponding to each element of the partition. Further, each element of the partition $\mathcal{A}_b$ is associated a parameter value, $\mu_{lb}$. Taken together the partition and the leaf parameters define a piecewise constant function: $g_l(x) = \mu_{lb}\ \text{if}\ x\in \mathcal{A}_b$; see Figure \ref{fig:treestep}. 

The tree prior $p(T_l)$ is specified by three components: (i) the probability of a node having children at
depth $d$
\begin{equation*}
\alpha (1+d)^{-\beta}, \qquad
\alpha \in (0,1), \beta \in [0, \infty),
\end{equation*}
(ii) the uniform distribution over available predictors for splitting rule assignment at
each interior node, and (iii) the uniform distribution on the discrete set of available splitting
values for the assigned predictor at each interior node.   This last choice has the appeal of invariance
under monotone transformations of the predictors.
 \cite{chipman2010bart} recommend $\alpha=.95$ and $\beta=2$ to enforce small trees. Finally, the leaf mean parameters, $\mu_{lb}$ are assigned independent mean-zero normal priors: $\mu_{lb}\sim \N(0,\tau)$. The parameter $\tau$ is a crucial regularization parameter; pointwise prior variance of $f$ is $\tau L$.

\subsection{The BART Splitting criterion}
By viewing the model as a data generating process, the Bayesian vantage point motivates modifications to the usual splitting criterion. Because the model stipulates that observations in the same leaf node share the same mean parameter, the prior predictive distribution --- obtained by integrating out the unknown group specific mean --- is simply a mean-zero multivariate normal distribution with covariance matrix 
\begin{equation*}
\mathbf{V} = \tau \mathrm{J}\mathrm{J}^t + \sigma^2 \mathbf{I},
\end{equation*}
where $\tau$ is the prior variance of the leaf-specific mean parameter, $\sigma^2$ is the variance of the additive error, and $\mathrm{J}$ is a column vector of all ones.  Observe that the prior predictive density of $\mathrm{y} \sim \N(0, \mathbf{V})$ is 
\begin{equation*}
p(\mathrm{y} \mid \tau, \sigma^2) = (2\pi)^{-n/2}\det(\mathbf{V})^{-1/2} \exp{\left( -\frac{1}{2} \mathrm{y}^t \mathbf{V}^{-1} \mathrm{y} \right)},
\end{equation*}
which can be simplified by a direct application of the matrix inversion lemma to $\mathbf{V}^{-1}$:
\begin{equation*}
\begin{aligned}
\mathbf{V}^{-1}&=  \sigma^{-2}\mathbf{I} - \frac{\tau}{\sigma^2(\sigma^2 + \tau n)} \mathrm{J}\mathrm{J}^t .
\end{aligned}
\end{equation*}
Applying Sylvester's determinant theorem to $\det{\mathbf{V}^{-1}}$  and taking logarithms yields
 a marginal log-likelihood of 
\begin{equation*}
\begin{aligned}
-\frac{n}{2}& \log{(2\pi)}-n\log{(\sigma)} + \frac{1}{2} \log{ \left( \frac{\sigma^2}{\sigma^2 + \tau n} \right)}   \\
&-\frac{1}{2} \frac{\mathrm{y}^t \mathrm{y}} {\sigma^{2}} + \frac{1}{2}\frac{\tau}{\sigma^2(\sigma^2 + \tau n)}s^2 ,
\end{aligned}
\end{equation*}
where we write $s \equiv \mathrm{y}^t\mathrm{J} = \sum_i y_i$ so that $\mathrm{y}^t  \mathrm{J}\mathrm{J}^t \mathrm{y} = (\sum_i y_i)^2 = s^2$. This likelihood is applied separately to partitions of the data corresponding to the leaves of a single fixed regression tree. Because observations in different leaf nodes are independent (conditional on $\sigma^2$), the full marginal log-likelihood is given by 
\begin{equation*}\label{loglikelihood}
\begin{aligned}
&\sum_{b=1}^B \left \lbrace -\frac{n_b}{2} \log{(2\pi)} -n_b\log{(\sigma)} +  \frac{1}{2} \log{ \left( \frac{\sigma^2}{\sigma^2 + \tau n_b} \right)} \right.\\
 & \hspace{0.3in} - \left. \frac{1}{2}\frac{\mathrm{y}_b^t \mathrm{y}_b} {\sigma^{2}} + \frac{1}{2}\frac{\tau}{\sigma^2(\sigma^2 + \tau n_b)} s_b^2 \right \rbrace \\
& = -n \log{(2\pi)} - n \log{(\sigma)} - \frac{1}{2} \frac{\mathrm{y}^t \mathrm{y}} {\sigma^{2}} \\
& \hspace{0.3in} + \frac{1}{2} \sum_{b=1}^B \left \lbrace  \log{ \left( \frac{\sigma^2}{\sigma^2 + \tau n_b} \right)}  + \frac{\tau}{\sigma^2(\sigma^2 + \tau n_b)} s_b^2 \right \rbrace
 \end{aligned}
 \end{equation*}
 where $b$ runs over all the leaf nodes and $\sum_{b=1}^B n_b = n$. Notice that the first three terms are not functions of the partition (the tree parameter), so they are constant, leaving 
 \begin{equation}\label{criterion}
\frac{1}{2} \sum_{b=1}^B \left \lbrace  \log{ \left( \frac{\sigma^2}{\sigma^2 + \tau n_b} \right)}  + \frac{\tau}{\sigma^2(\sigma^2 + \tau n_b)} s_b^2 \right \rbrace
\end{equation}
as the model-based split criterion, where $(n_b, s_b, B)$ are functions of the data and the tree $T$.

\subsection{The BART MCMC}\label{bart}
The basic BART MCMC proceeds as a Metropolis-within-Gibbs algorithm, with the key update of the individual regression trees being conducted as a  local random walk Metropolis-Hastings (MH) update, given all of the other trees as well as the residual variance parameter, $\sigma^2$. Let $\mathcal{T}$ denote the set of trees and $\mathcal{M}$ denote the set of leaf parameter vectors. Recall that $|\mathcal{T}| = |\mathcal{M}| = L$, and each $\mu_l \in \mathcal{M}$ is length $B(l)$. 

The sequence of Gibbs updates are
\begin{enumerate}
\item $T_l, \mu_l \mid \mathcal{T}_{-l}, \mathcal{M}_{-l}, \sigma^2, \y$, for $l = 1, \dots, L$, which is done compositionally (for each $l$) as
\begin{enumerate}
\item $T_l \mid \mathcal{T}_{-l}, \mathcal{M}_{-l}, \sigma^2, \y$,
\item $\mu_l \mid \mathcal{T}, \mathcal{M}_{-l}, \sigma^2, \y$,
\end{enumerate}
\item $\sigma^2 \mid \mathcal{T}, \mathcal{M}, \y$.
\end{enumerate}
Taking advantage of the additive structure of the model, these updates can be written as
\begin{enumerate}
\item $T_l, \mu_l \mid \res_l, \sigma^2$, for $l = 1, \dots, L$, which is done compositionally (for each $l$) as
\begin{enumerate}
\item $T_l \mid \res_l, \sigma^2$,
\item $\mu_l \mid T_l, \res_l, \sigma^2$,
\end{enumerate}
\item $\sigma^2 \mid \res$.
\end{enumerate}
for ``residuals'' defined as
\begin{equation*}
\res_l^{(k+1)} \equiv \y - \sum_{l' < l} g(\X;T_{l'},\mu_{l'})^{(k+1)} - \sum_{l' > l} g(\X;T_{l'},\mu_{l'})^{(k)},
\end{equation*}
and
\begin{equation*}
\res^{(k)} \equiv \y - \sum_{l = 1}^L g(\X;T_{l},\mu_{l})^{(k)},
\end{equation*}
where $k$ indexes the Monte Carlo iteration. Update 1(a) is a Metropolis-Hastings update based on the integrated likelihood given in (\ref{criterion}). Update 1(b) is a conditionally conjugate Gaussian mean update done separately for each leaf node parameter $\mu_{lb}$, $b = 1\dots B(l)$. Update 2 is a conditionally conjugate inverse-Gamma update. 

Step 1(a) is handled with a random walk as follows. Given a current tree, $T$, modifications are proposed and
either accepted or rejected according to a likelihood ratio based on (\ref{criterion}).  \cite{chipman1998bayesian} describes proposals comprising a birth/death pair, in which a birth spawns to children from a given bottom node
and a death kills a pair of sibling children; see \cite{pratola2016efficient} for alternative choices. For example, in a birth move, a variable to split on, as well as a cut-point to split at, are selected uniformly at random from the available splitting rules. Via these simple MH updates, BART stochastically searches through regression models of varying complexity (in terms of tree-depth). For ``smaller'' problems, with dozens of predictors and thousands of observations, this MCMC approach has proven to be remarkably effective; for larger problems, with hundreds of thousands of observations, it does not work well on standard desktops.

In the next section, we present our new stochastic hill climbing algorithm called accelerated Bayesian additive regression trees (XBART), see algorithm \ref{alg:XBART}. It follows the Gibbs update framework but replace the Metropolis-Hastings updates of each single tree by a new grow-from-root backfitting strategy; see Algorithm \ref{alg:tree}.


\section{XBART}
\subsection{Grow-from-root backfitting} 
Rather than making small moves to a given tree $T_l^{(k)}$ at iteration $k+1$, here we ignore the current tree and grow an entirely new tree $T_l^{(k+1)}$ from scratch. We grow each tree recursively and stochastically and the tree growing process is also terminated stochastically, based on the ``residual'' data defined above. The pseudo-code is presented in Algorithm \ref{alg:tree}.

Specifically, at each level of the recursion we consider every available cut-point (decision rule threshold) for each variable\footnote{For simplicity, in this paper we consider only continuous predictor variables.} and evaluate the integrated likelihood criterion, the exponential of expression (\ref{criterion}). We also consider the no-split option, which corresponds to a cut-point outside of the range of the available data. How many such {\em null} cut-points to consider is a modeling decision; we default to one such null cut-point per variable. Accordingly, with $C$ available active cut-points and $V$ total variables we perform $C\times V + 1$ likelihood evaluations. Each of the active cut-points is weighted by $\alpha (1+d)^{-\beta}$ and the unweighted cut-points weighted by $1 - \alpha (1+d)^{-\beta}$, as per the prior\footnote{Equivalently, the active cut-points are equally weighted and the no split option is weighted $V (\alpha^{-1}(1+d)^{\beta} - 1)$. An additional multiplier could be used here to encourage/discourage tree growth.}. Since data is pre-sorted, we index candidate cut-points by their rank, $c = 0, 1, \cdots, C\times V $ and $c = 0$ denotes a {\em null} cut-point, the ``do not split'' option. Selection of a variable to split on, and a cut-point to split at,are then chosen by Bayes rule:
\begin{equation}\label{sample}
\pi(v, c) = \frac{\exp{(\ell(c, v)}) \kappa(c)}{\sum_{v' = 1}^V \sum_{c'=0}^C  \exp{(\ell(c',v'))} \kappa(c')}
\end{equation}
where 
\small
\begin{equation*}
\begin{aligned}
\ell(v,c) &= \frac{1}{2} \left \lbrace  \log{ \left( \frac{\sigma^2}{\sigma^2 + \tau n(\leq,v,c)} \right)}  \right.\\
& \hspace{0.3in}+\left. \frac{\tau}{\sigma^2(\sigma^2 + \tau n(\leq,v,c))} s(\leq,v,c)^2 \right \rbrace\\
& + \frac{1}{2} \left \lbrace  \log{ \left( \frac{\sigma^2}{\sigma^2 + \tau n(>,v,c)} \right)}  \right.\\
& \hspace{0.3in} +\left. \frac{\tau}{\sigma^2(\sigma^2 + \tau n(>,v,c))} s(>,v,c)^2 \right \rbrace
 \end{aligned}
 \end{equation*}
\normalsize
for $c \neq 0$. Here $n(\leq,v,c)$ is the number of observations in the current leaf node that have $x_v \leq c$ and $s(\leq, v, c)$ is the sum of the residual $\res_l^{(k)}$ of those same observations; $n(>, v,c)$ and $s(>, v,c)$ are defined analogously.  Also, $\kappa(c \neq 0) = 1$. 

For $c = 0$, corresponding to null cut-points or the stop-splitting option, we have instead 
\small
\begin{equation*}
\ell(v,c) = \frac{1}{2} \left \lbrace  \log{ \left( \frac{\sigma^2}{\sigma^2 + \tau n} \right)}  + \frac{\tau}{\sigma^2(\sigma^2 + \tau n)} s^2 \right \rbrace
 \end{equation*}
\normalsize
and $\kappa(0) = \frac{1 - \alpha (1+d)^{-\beta}}{\alpha (1+d)^{-\beta}}$, where $n$ denotes the number of observations in the current leaf node, $n = n(\leq,v,c) + n(>,v,c)$ and $s$ denotes the sum over all the current leaf data.

Using this new tree-growing strategy, we find that different default parameters are advisable. We recommend $L = \frac{1}{4} (\log{n})^{\log \log n}$, $\alpha = 0.95$, $\beta = 1.25$ and $\tau = \frac{3}{10} \mbox{var}(\y)/L$. This choice of $L$ is a function that is faster growing than $\log{n}$, but slower than $\sqrt{n}$, while the lower value of $\beta$ permits deeper trees (than BART's default $\beta = 2$). Allowing $L$ to grow as a function of the data permits smoother functions to be estimated more accurately as the sample size grows, whereas a sample size-independent choice would be limited in its smoothness by the number of trees. The suggested choice of $\tau$ dictates that {\em a priori} the function will account for 30\% of the observed variance of the response variable. Finally, while BART must be run for many thousands of iterations with a substantial burn-in period, our default suggestion is just 40 sweeps through the data, discarding the first 15 as burn-in.

\begin{algorithm*}[h]
\small
\caption{Grow-from-root backfitting}\label{alg:tree}
\begin{algorithmic}
\Procedure{grow\_from\_root}{$\y$, $\X$, $C$, $m$, $\w$, $\sigma^2$}\Comment{Fit a tree using data $y$ and $\X$ by recursion.}\\
\textbf{output} A tree $T_l$ and a vector of split counts $\w_l$.
\State $N\gets $ number of rows of $y, x$
\State Sample $m$ variables use weight $w$ as shown in section \ref{mtry}. 
\State Select $C$ cutpoints as shown in section \ref{cutpoints}.
\State Evaluate $C\times m + 1$ candidate cutpoints and no-split option with equation (\ref{sample}).
\State Sample one cutpoint propotional to equation (\ref{sample}).
\If{sample no-split option} 
\State Sample leaf parameter from normal distribution $\mu\sim N\left(\sum y/\left[\sigma^2\left(\frac{1}{\tau} + \frac{N}{\sigma^2}\right)\right], 1/\left[\frac{1}{\tau} + \frac{N}{\sigma^2}\right]\right)$. \textbf{return}
\Else 
\State $w_l[j] = w_l[j] + 1$, add count of selected split variable.
\State Split data to left and right node.
\State GROW\_FROM\_ROOT($y_{\text{left}}$,$\X_{\text{left}}$, $C$, $m$, $\w$, $\sigma^2$)
\State GROW\_FROM\_ROOT($y_{\text{right}}$,$\X_{\text{right}}$, $C$, $m$, $\w$, $\sigma^2$)
\EndIf
\EndProcedure
\end{algorithmic}
\end{algorithm*}

\subsection{Pre-sorting Features for Efficiency}
Observe that the BART criterion depends on the partition sums only. An important implication of this, for computation, is that with sorted predictor variables the various cut-point integrated likelihoods can be computed rapidly via a single sweep through the data (per variable), taking cumulative sums. Let $\mathbf{O}$ denote the $V$-by-$n$ array such that $o_{vh}$ denotes the index, in the data, of the observation with the $h$th smallest value of the $v$th predictor variable $x_v$. Then, taking the cumulative sums gives
\begin{equation}
s(\leq,v,c) = \sum_{h \leq c} \res_{o_{vh}}
 \end{equation}
 and 
 \begin{equation}
s(>,v,c) = \sum_{h = 1}^n r_{lh}-  s(\leq,v,c).
\end{equation}
The subscript $l$ on the residual indicates that these evaluations pertain to the update of the $l$th tree. 

The above formulation is useful if the data can be presorted and, furthermore, the sorting can be maintained at all levels of the recursive tree-growing process. To achieve this, we must ``sift'' each of the variables before passing to the next level of the recursion. Specifically, we form two new index matrices $\mathbf{O}^{\leq}$ and $\mathbf{O}^{>}$ that partition the data according to the selected split rule. For the selected split variable $v$ and selected split $c$, this is automatic: $O_v^{\leq} = O_{v,1:c}$ and $O_v^{>} = O_{v,(c+1):n}$. For the other $V-1$ variables, we sift them by looping through all $n$ available observations, populating $O^{
\leq}_{q}$ and $O^{>}_{q}$, for $q \neq v$, sequentially, with values $o_{qj}$ according to whether $x_{vo_{qj}} \leq c$ or $x_{vo_{qj}} > c$, for $j = 1, \dots, n$.

Because the data is processed in sorted order, the ordering will be preserved in each of the new matrices $\mathbf{O}^{\leq}$ and $\mathbf{O}^{>}$. This strategy was first presented in \cite{mehta1996sliq} in the context of tree classification algorithms.
\vspace{-3mm}

\subsection{Recursively Defined Cut-points}\label{cutpoints}
Evaluating the integrated likelihood criterion is straightforward, but the summation and normalization required to sample the cut-points contribute a substantial computational burden in its own right. Therefore, it is helpful to consider a restricted number of cut-points $C$. 
This can simply be achieved  by taking every $j$th value (starting from the smallest) as an eligible split point with $j = \lfloor \frac{n_b-2}{C} \rfloor$. As the tree grows deeper, the amount of data that is skipped over diminishes.  Eventually we get $n_b < C$, and each data point defines a unique cut-point. In this way the data could, without regularization, be fit perfectly, even though the number of cut-points at any given level is given an upper limit. As a default, we set the number of cut-points to $\max{(\sqrt{n},100)}$, where $n$ is the sample size of the entire data set. 

Our cut-point subsampling strategy is more naive than the cut-point subselection search heuristics used by  {\tt XGBoost} \citep{chen2016xgboost} and {\tt LightGBM} \citep{ke2017lightgbm}, which both consider the gradient evaluated at each cut-point when determining the next split. Our approach does not consider the response information at all, but rather defines a predictor-dependent prior on the response surface. That is, given a design matrix $\mathbf{X}$, a sample functions can be drawn from the prior distribution by sampling trees, splitting uniformly at random among the cut-points defined by the node-specific quantiles, in a sequential fashion. In further contrast, 
the proposed method stochastically samples cut-points proportional to its objective function, rather than deterministically maximizing the likelihood-prior. Then, multiple sweeps are made through the data. 
Rather than greedy (approximate) optimization, like {\tt XGBoost} and {\tt LightGBM}, the proposed algorithm performs a stochastic hill climb by coordinate ascent over multiple sweeps through the parameters.  

\subsection{Sparse Proposal Distribution}\label{mtry}
As a final modification, we strike an intermediate balance between the local BART updates, which randomly consider one variable at a time, and the all-variables Bayes rule described above. We do this by considering $m \leq V$ variables at a time when sampling each splitting rule. Rather than drawing these variables uniformly at random, as done in random forests, we introduce a parameter vector $\w$ which denotes the prior probability that a given variable is chosen to be split on, as suggested in \cite{linero2016bayesian}. Before sampling each splitting rule, we randomly select $m$ variables with probability proportional to $\w$. These $m$ variables are sampled sequentially and {\em without replacement}, with selection probability proportional to $\w$.

The variable weight parameter $\w$ is given a Dirichlet prior with hyperparameter $\bar{\w}$ set to all ones and subsequently incremented to count the total number of splits across all trees. The split counts are then updated in between each tree sampling/growth step: 
\begin{equation}
\bar{\w} \leftarrow \bar{\w} - \bar{\w}_l^{(k-1)} +\bar{\w}_l^{(k)}
\end{equation}
 where $\bar{\w}_l^{(k)}$ denotes the length-$V$ vector recording the number of splits on each variable in tree $l$ at iteration $k$. The weight parameter is then resampled as
$\w \sim \mbox{Dirichlet}(\bar{\w}).$ 
Splits that improve the likelihood function will be chosen more often than those that don't. The parameter $\w$ is then updated to reflect that, making chosen variables more likely to be considered in  subsequent sweeps. In practice, we find it is helpful to use all $V$ variables during an initialization phase, to more rapidly obtain an accurate initial estimate of $\w$.  

\subsection{The Estimator}
Given $K$ iterations of the algorithm, the final $K -  I$ samples are used to compute a point-wise average function evaluation, where $I < K$ is denotes the length of the burn-in period. As mentioned above, we recommend $K = 40$ and $I = 15$ for routine use. The final estimator is therefore expressible as 
\begin{equation}
\bar{f}(\X) = \frac{1}{K-I}\sum_{k > I}^K f^{(k)}(\X).
\end{equation}
where $f^{(k)}$ denotes a sample of the forest, as in expression \ref{forest}, drawn by algorithm \ref{alg:XBART}. We note that this corresponds to the Bayes optimal estimator under mean squared error estimation loss, provided that we have samples from a legitimate posterior distribution. As the grow-from-root strategy is not a proper full conditional, this estimator must be considered a greedy stochastic approximation (but see section \ref{mh}). Nonetheless, simulation results strongly suggest that the approximation is adequate. 

A few remarks on posterior uncertainty. First, with only $K = 40$ sweeps, the XBART posterior uncertainty is likely understated. However, the standard BART MCMC is probably not mixing well in most contexts, either, and yet still provides useful, if approximate, uncertainty quantification. Second, experiments with a version of XBART based on only the final sweep, $K - I = 1$, performed worse than methods with $K-I > 1$, suggesting that our posterior exploration, while imperfect, is still beneficial.

\begin{algorithm*}[h]
\small
\caption{Accelerated Bayesian Additive Regression Trees (XBART)}\label{alg:XBART}
\begin{algorithmic}
\Procedure{XBART}{$\y, \X,C,m, L, I,  K, \alpha, \eta$} \Comment($\alpha, \eta$ are prior parameter of $\sigma^2$)\\
\textbf{output} Samples of forest
\State $V\gets $ number of columns of $\X$
\State $N\gets $ number of rows of $\X$
\State Initialize $\res_l^{(0)} \leftarrow \y / L$.
\For{$k$ in 1 to $K$}
\For{$l$ in 1 to $L$}
\State Calculate residual $\res_l^{(k)}$ as shown in section \ref{bart}. 
\If{$k < I $}
\State  GROW\_FROM\_ROOT($\res_l^{(k)}$,$\X$, $C$, $V$, $\w$, $\sigma^2$) \Comment{use all variables in burnin iterations}
\Else 
\State  GROW\_FROM\_ROOT($\res_l^{(k)}$,$\X$, $C$, $m$, $\w$, $\sigma^2$)
\EndIf
\State $\bar{\w} \gets \bar{\w} - \bar{\w}_l^{(k-1)}+ \bar{\w}_l^{k}$ \Comment{update $\bar{\w}$ with split counts of current tree}
\State $\w \sim \mbox{Dirichlet}(\bar{\w})$ 
\State $\sigma^2 \sim \mbox{Inverse-Gamma}(N + \alpha, \res_l^{(k)t}\res_l^{(k)} + \eta)$ 
\EndFor
\EndFor
\textbf{return}
\EndProcedure
\end{algorithmic}
\end{algorithm*}

\subsection{Metropolis-Hastings Proposal Distribution}\label{mh}
A fully Bayesian algorithm can be obtained by using the grow-from-root fitting algorithm as a data-driven Metropolis-Hastings proposal distribution. Importantly, the MH accept-reject step should be completed at the end of each {\em sweep}, that is, after proposing an entirely new set of trees and their associated parameters. Denote the current and proposed sets, repectively, by $F = \{\mathcal{T}, \mathcal{M}\}$ and $F' = \{\mathcal{T}', \mathcal{M}'\}$, where $\mathcal{T} = \{T_1, T_2, \dots, T_L\}$ and $\mathcal{M} = \{\mu_1, \mu_2, \dots, \mu_L\}$ denote the set of trees and leaf parameters, respectively. The grow-from-root algorithm generates a proposal of moving from $F$ to $F'$ with density $q(F', F)$ defined by a recursive product of terms as in 3.1. The probability of growing any particular tree is characterized by the probability of a certain sequence of split (or no-split) decisions encountered as one navigates down a given tree. The density of the leaf parameters, conditional on a given tree structure, follows from the corresponding conjugate normal update. See Algorithm \ref{alg:growprob}. To show that this MH procedure is valid, we need only show that any set of trees and parameters can be reached from any other set (positive recurrence) and that the proposal density is well-defined upon interchanging the sets of tree/parameter pairs; the construction of the usual Metropolis-Hastings ratio ensures detailed balance. Observe that one initializes the proposal process starting from a residual vector defined by $F$. To propose the first tree in $F'$, we ``kill'' the first tree from $F$ and grow an entirely new tree. In the second step, we recompute the residual and repeat, and so forth. After $L$ steps, $L$ new trees have been regrown in an unrestricted fashion. Although the trees grown in this sequence are not independent, their joint density is given by a product of conditional densities, all of the dependence being passed through the redefinition of the residual at each step; see Algorithm \ref{alg:evalprop}. Consequently, one can interchange the roles of $F$ and $F'$ in this elaborate proposal mechanism simply by beginning the process with the residual defined by $F'$ rather than $F$. Further work will consider the efficacy of this approach.


%
%
%

\vspace{-3mm}

\section{SIMULATION STUDIES}

\subsection{Data Generating Process}
To demonstrate the performance of the new accelerated BART heuristic, which we call XBART, we estimate function evaluations with a hold-out set that is a quarter of the training sample size and judge accuracy according to root mean squared error (RMSE). We consider four different challenging functions, $f$, as defined in Table \ref{tab:truef}. In all cases, $x_j \iid \N(0,1)$  for $j = 1, \dots, d = 30$. The data is generated according to the additive error mode (\ref{additive}), with $\epsilon_i \iid \N(0,1)$. We consider $\sigma = \kappa \mbox{Var}(f)$ for $\kappa \in \lbrace 1, 10 \rbrace$.

\subsection{Methods}
We compare to leading machine learning algorithms: random forests, gradient boosting machines, neural networks, and BART MCMC. All implementations had an {\tt R} interface and were the current fastest implementations to our knowledge: {\tt ranger} \citep{wright2015ranger}, {\tt xgboost} \citep{chen2016xgboost}, and {\tt Keras} \citep{chollet2015keras}, {\tt dbarts} respectively. For {\tt Keras} we used a single strong architecture but varied epochs depending on the noise in the problem. For {\tt xgboost} we consider two specifications, one using the software defaults and another determined by by 5-fold cross-validated grid optimization (see Table \ref{tab:f7}); a reduced grid of parameter values was used at sample sizes $n > 10,000$. Comparison with  {\tt ranger} and {\tt dbarts} are shown in supplementary material.
\begin{algorithm}[h]
\caption{Grow Probability}\label{alg:growprob}
\small
\begin{algorithmic}
\Procedure{GrowProb}{$r, T, \mu, X, h$} 
\State $\psi_h \leftarrow \pi(v_h(T), c_h(T)$ \Comment{From equation (3)}
\If{$v_h(T)$ = NULL} \Comment{If this is bottom node}
\State $\psi_h \leftarrow \psi_h \times \phi(\mu_h \mid \mu, \sigma^2)$
\Else
\State$\psi_h \leftarrow  $GrowProb($r_\text{left}, T, \mu, 2h$)
\State$\psi_h \leftarrow  $GrowProb($r_\text{right}, T, \mu, 2h+1$)
\EndIf
\State\textbf{return} $\psi_h$
\EndProcedure
\end{algorithmic}
\end{algorithm}

\vspace{-3mm}

\begin{algorithm}[h]
\small
\caption{Evaluate Proposal Density}\label{alg:evalprop}
\begin{algorithmic}
\Procedure{PropDens}{$F, F', y, \sigma^2, \tau, x$} 
\State Construct residual $r \leftarrow y - f(F_{2:L})$, initialize $q \leftarrow 1$
\For{$l$ in 1 to $L$}
\State Set $\psi$$\leftarrow $Prod(GROWPROB($r, F'_l, \mu_l, x, h = 1$))
\State $q \leftarrow q\times\psi$ 
\State Update residual $r \leftarrow y - f(F_{(l+1):L}) - f(F'_{1:l})$
\EndFor
\State\textbf{return} $q = q(F', F)$
\EndProcedure
\end{algorithmic}
\end{algorithm}
\subsection{Computation}

The software used was \texttt{R} version 3.4.4 with  \texttt{xgboost} 0.71.2, \texttt{dbarts} version 0.9.1, \texttt{ranger} 0.10.1 and \texttt{keras} 2.2.0. The default hyperparameters for XGBoost  are \texttt{eta} $= 0.3$, {\tt colsample\_bytree} $=1$, {\tt min\_child\_weight} $= 1$ and \texttt{max\_depth} $= 6$. Ranger was fit with \texttt{num.trees} $=500$ and \texttt{mtry} $ = 5 \approx \sqrt{d}$. BART, with the package {\tt dbarts}, was fit with the defaults of {\tt ntrees} $= 200$, {\tt alpha} $= 0.95$, {\tt beta} $=2$, with a burn-in of 5,000 samples ({\tt nskip} $=5000$) and 2,000 retrained posterior samples ({\tt ndpost} $=2000$). 
\vspace{-3mm}
\begin{table}[h]
\small
\def\arraystretch{1}
\centering 
\caption{Four true $f$ functions} 
\begin{tabular}{l|m{2.15in}} 
\toprule
Name &Function \\
\hline
Linear &     $ \x^t \mathrm{\gamma} $;\; $ \gamma_j = -2+ \frac{4(j-1)}{d-1} $\\ 

Single index & $10\sqrt{a} + \sin{(5a)}$;\;$a=\sum_{j=1}^{10} (x_j - \gamma_j)^2$;\; $\gamma_j = -1.5+ \frac{j-1}{3}$.\\


Trig + poly & $5\sin(3x_1)+2x_2^2 + 3x_3x_4 $\\

Max & $\max(x_1,x_2,x_3) $\\ 
\bottomrule
\end{tabular}
\label{tab:truef}
\end{table}
The default {\tt dbarts} algorithm uses an evenly spaced grid of 100 cut-point candidates along the observed range of each variable ({\tt numcuts} $=100$, {\tt usequants = FALSE}). For {\tt Keras} we build a network with two hidden layers (15 nodes each) using ReLU activation function, $\ell_1$ regularization at 0.01, and with 50/20 epochs depending on the signal to noise ratio. 
\vspace{-3mm}

 \begin{table}[h]
  \centering
  \small
        \caption{Hyperparameter Grid for XGBoost}
    \begin{tabular}{l|rr}
\toprule
     Parameter name     & $N = 10$K & $N > 10$K \\
          \hline 
    {\tt eta}   & $\lbrace 0.1, 0.3\rbrace$ &$\lbrace 0.1, 0.3\rbrace$ \\
    
    {\tt max\_depth} &$\lbrace 4, 8, 12\rbrace$ & $\lbrace 4, 12\rbrace$ \\ 
    {\tt colsample\_bytree} & $\lbrace 0.7, 1 \rbrace$  & $\lbrace 0.7, 1 \rbrace$ \\ 
    {\tt min\_child\_weight} & $\lbrace 1, 10, 15 \rbrace$ & $10$ \\ 
    {\tt subsample} & 0.8   & 0.8 \\ 
    {\tt gamma }& 0.1   & 0.1  \\
    \bottomrule
    \end{tabular}%
  \label{tab:f7}%
\end{table}%
\vspace{-3mm}
\subsection{Results}

The performance of the new XBART algorithm was excellent, showing superior speed and performance relative to all the considered alternatives on essentially every data generating processes. The full results, averaged across five Monte Carlo replications, are reported in Tables \ref{tab:allall}. Neural networks perform as well as XBART in the low noise settings under the Max and Linear functions. Unsurprisingly, neural networks outperform XBART under the linear function with low noise. Across all data generating processes and sample sizes, XBART was 31\% more accurate than the cross-validated XGBoost method and typically faster. Specifically, the supplement examines the empirical examples given in \cite{chipman2010bart}.

The XBART method was slower than the untuned default XGBoost method, but was 3.5 times more accurate. This pattern points to one of the main benefits of the proposed method, which is that it has excellent performance using the same hyperparameter settings across all data generating processes. Importantly, these default hyperparameter settings were decided on the basis of prior elicitation experiments using different true functions than were used in the reported simulations. While XGBoost  is quite fast, the tuning processes is left to the user and can increase the total computational burden by orders of magnitude.

Random forests and traditional MCMC BART were prohibitively slow at larger sample sizes. However, at $n = 10,000$ several notable patterns did emerge; see the supplementary material for full details. First was that BART and XBART typically gave very similar results, as would be expected. BART performed slightly better in the low noise setting and quite a bit worse in the high noise setting (likely due to inadequate burn-in period). Similarly, random forests do well in higher noise settings, while XGBoost and neural networks perform better in lower noise settings. 

%
%
%
%
%

\section{DISCUSSION}
The grow-from-root strategy proposed here opens the door for computational innovations to be married to the novel BART stochastic fitting algorithm. Further, the proposed adaptive cut-points and variable selection proposal together define a novel predictor-dependent prior, marking a distinct Bayesian model. The simulation studies clearly demonstrate the beneficial synergy realized by the proposed approach: XBART is a state-of-the-art nonlinear regression method with computational demands that are competitive with the current fastest alternatives. In particular, the excellent performance without the need to cross-validate recommends XBART as a suitable default method for function estimation and prediction tasks when little is known about the response surface.

\addtolength{\tabcolsep}{-2pt}
\begin{table}[h]
  \centering
  \scriptsize
    \begin{tabular}{rllll}
          \multicolumn{5}{c}{$\kappa = 1$} \\\hline
    \toprule
$n$ & XBART    & XGB+CV    & XGB   & NN \\
    \toprule
    \multicolumn{5}{c}{Linear} \\\hline
     10k & 1.74 (20) & 2.63 (64) & 3.23 (0) & 1.39 (26) \\
     50k & 1.04 (180) & 1.99 (142) & 2.56 (4) & 0.66 (28) \\
     250k& 0.67 (1774) & 1.50 (1399) & 2.00 (55) & 0.28 (40) \\
    \toprule
    \multicolumn{5}{c}{Max} \\\hline
     10k & 0.39 (16) & 0.42 (62) & 0.79 (0) & 0.40 (30) \\
     50k & 0.25 (134) & 0.29 (140) & 0.58 (4) & 0.20 (32) \\
     250k & 0.14 (1188) & 0.21 (1554) & 0.41 (60) & 0.16 (44) \\
    \toprule
    \multicolumn{5}{c}{Single Index} \\\hline
    10k  & 2.27 (17) & 2.65 (61) & 3.65 (0) & 2.76 (28) \\
    50k & 1.54 (153) & 1.61 (141) & 2.81 (4) & 1.93 (31) \\
    250k & 1.14 (1484) & 1.18 (1424) & 2.16 (55) & 1.67 (41) \\
    \toprule
    \multicolumn{5}{c}{Trig + Poly} \\\hline
    10k  & 1.31 (17) & 2.08 (61) & 2.70 (0) & 3.96 (26) \\
    50k & 0.74 (147) & 1.29 (141) & 1.67 (4) & 3.33 (29) \\
    250k & 0.45 (1324) & 0.82 (1474) & 1.11 (59) & 2.56 (41) \\
    \toprule

\\
      \multicolumn{5}{c}{$\kappa = 10$} \\\hline

    \toprule
     $n$ & XBART    & XGB+CV    & XGB   & NN \\
    \toprule
    \multicolumn{5}{c}{Linear} \\\hline
     10k  & 5.07 (16) & 8.04 (61) & 21.25 (0) & 7.39 (12) \\
     50k & 3.16 (135) & 5.47 (140) & 16.17 (4) & 3.62 (14) \\
     250k & 2.03 (1228) & 3.15 (1473) & 11.49 (54) & 1.89 (19) \\
    \toprule
    \multicolumn{5}{c}{Max} \\\hline
     10k & 1.94 (16) & 2.76 (60) & 7.18 (0) & 2.98 (15) \\
     50k & 1.22 (133) & 1.85 (139) & 5.49 (4) & 1.63 (16) \\
    250k & 0.75 (1196) & 1.05 (1485) & 3.85 (54) & 0.85 (22) \\
    \toprule
    \multicolumn{5}{c}{Single Index} \\\hline
     10k & 7.13 (16) & 10.61 (61) & 28.68 (0) & 9.43 (14) \\
    50k & 4.51 (133) & 6.91 (139) & 21.18 (4) & 6.42 (16) \\
    250k & 3.06 (1214) & 4.10 (1547) & 14.82 (54) & 4.72 (21) \\
    \toprule
    \multicolumn{5}{c}{Trig + Poly} \\\hline
     10k & 4.94 (16) & 7.16 (61) & 17.97 (0) & 8.20 (13) \\
    50k & 3.01 (132) & 4.92 (139) & 13.30 (4) & 5.53 (14) \\
    250k & 1.87 (1216) & 3.17 (1462) & 9.37 (49) & 4.13 (20) \\
    \bottomrule
    \end{tabular}%
 \caption{Root mean squared error (RMSE) of each method. Column XGB+CV is result of XGBoost with tuning parameter by cross validation. The number in parenthesis is running time in seconds. First column is number of data observations (in thousands).}
  \label{tab:allall}%
\end{table}%
The source of XBART's superior performance is not entirely clear, but preliminary investigations point to two important factors. One, the BART splitting criterion involves (the current estimate of) the error standard deviation, $\sigma$, meaning that it is adaptively regularizing within the model fitting process. Two, we conjecture that the stochastic nature of the algorithm leads to better exploration of the parameter space than iterative optimizers. With fast model fitting software now in hand, this issue can be investigated more systematically in future work. Another line of future research is to incorporate XBART within extended BART models such as Bayesian causal forests \citep{hahn2017bayesian} and BART for log-linear models \citep{murray2017log}.

\clearpage
\bibliography{BART}

\begin{thebibliography}{}

\bibitem[Breiman, 1997]{breiman1997arcing}
Breiman, L. (1997).
\newblock Arcing the edge.
\newblock Technical report, Technical Report 486, Statistics Department,
  University of California at Berkeley.

\bibitem[Breiman, 2001]{breiman2001random}
Breiman, L. (2001).
\newblock Random forests.
\newblock {\em Machine learning}, 45(1):5--32.

\bibitem[Breiman et~al., 1984]{breiman1984classification}
Breiman, L., Friedman, J., Olshen, R., and Stone, C.~J. (1984).
\newblock {\em Classification and regression trees}.
\newblock Chapman and Hall/CRC.

\bibitem[Chen and Guestrin, 2016]{chen2016xgboost}
Chen, T. and Guestrin, C. (2016).
\newblock {XGB}oost: A scalable tree boosting system.
\newblock In {\em Proceedings of the 22nd ACM SIGKDD International Conference
  on Knowledge Discovery and Data Mining}, pages 785--794. ACM.

\bibitem[Chipman et~al., 1998]{chipman1998bayesian}
Chipman, H.~A., George, E.~I., and McCulloch, R.~E. (1998).
\newblock Bayesian {C}{A}{R}{T} model search.
\newblock {\em Journal of the American Statistical Association},
  93(443):935--948.

\bibitem[Chipman et~al., 2010]{chipman2010bart}
Chipman, H.~A., George, E.~I., McCulloch, R.~E., et~al. (2010).
\newblock {BART}: Bayesian additive regression trees.
\newblock {\em The Annals of Applied Statistics}, 4(1):266--298.

\bibitem[Chollet et~al., 2015]{chollet2015keras}
Chollet, F. et~al. (2015).
\newblock Keras.

\bibitem[Friedman, 2001]{friedman2001greedy}
Friedman, J.~H. (2001).
\newblock Greedy function approximation: a gradient boosting machine.
\newblock {\em Annals of Statistics}, pages 1189--1232.

\bibitem[Friedman, 2002]{friedman2002stochastic}
Friedman, J.~H. (2002).
\newblock Stochastic gradient boosting.
\newblock {\em Computational Statistics \& Data Analysis}, 38(4):367--378.

\bibitem[Hahn et~al., 2017]{hahn2017bayesian}
Hahn, P.~R., Murray, J.~S., and Carvalho, C. (2017).
\newblock Bayesian regression tree models for causal inference: regularization,
  confounding, and heterogeneous effects.
\newblock {\em arXiv preprint arXiv:1706.09523}.

\bibitem[Ke et~al., 2017]{ke2017lightgbm}
Ke, G., Meng, Q., Finley, T., Wang, T., Chen, W., Ma, W., Ye, Q., and Liu,
  T.-Y. (2017).
\newblock {LightGBM}: A highly efficient gradient boosting decision tree.
\newblock In {\em Advances in Neural Information Processing Systems}, pages
  3146--3154.

\bibitem[Linero, 2016]{linero2016bayesian}
Linero, A.~R. (2016).
\newblock Bayesian regression trees for high dimensional prediction and
  variable selection.
\newblock {\em Journal of the American Statistical Association},
  (just-accepted).

\bibitem[Mehta et~al., 1996]{mehta1996sliq}
Mehta, M., Agrawal, R., and Rissanen, J. (1996).
\newblock {SLIQ}: A fast scalable classifier for data mining.
\newblock In {\em International Conference on Extending Database Technology},
  pages 18--32. Springer.

\bibitem[Murray, 2017]{murray2017log}
Murray, J.~S. (2017).
\newblock Log-linear bayesian additive regression trees for categorical and
  count responses.
\newblock {\em arXiv preprint arXiv:1701.01503}.

\bibitem[Pratola, 2016]{pratola2016efficient}
Pratola, M. (2016).
\newblock Efficent {M}etropolis-{H}astings proposal mechanism for {B}ayesian
  regression tree models.
\newblock {\em Bayesian Analysis}, 11(3):885--911.

\bibitem[Wright and Ziegler, 2015]{wright2015ranger}
Wright, M.~N. and Ziegler, A. (2015).
\newblock ranger: {A} fast implementation of random forests for high
  dimensional data in {C}++ and {R}.
\newblock {\em arXiv preprint arXiv:1508.04409}.

\end{thebibliography}

\end{document}